\documentclass{article}

\usepackage{arxiv}

\usepackage[utf8]{inputenc} 
\usepackage[T1]{fontenc}    
\usepackage{hyperref}       
\usepackage{url}            
\usepackage{booktabs}       
\usepackage{amsfonts}       
\usepackage{nicefrac}       
\usepackage{microtype}      
\usepackage{lipsum}
\usepackage[export]{adjustbox}

\usepackage[section]{placeins}
\usepackage[shortlabels]{enumitem}
\usepackage{subcaption}
\usepackage[numbers]{natbib}
\usepackage{amsmath,amssymb,amsfonts}
\usepackage{algorithmic}
\usepackage{graphicx}
\usepackage{textcomp}
\usepackage{xcolor}
\def\BibTeX{{\rm B\kern-.05em{\sc i\kern-.025em b}\kern-.08em
    T\kern-.1667em\lower.7ex\hbox{E}\kern-.125emX}}
    
\usepackage{verbatim}  

\usepackage{hyperref}
\usepackage{csquotes}
\usepackage{todonotes}
\usepackage{float}

\usepackage{diagbox}
\usepackage{multirow}
\usepackage{tabularx}
\newcolumntype{Y}{>{\centering\arraybackslash}X}

\usepackage{ tipa }

\usepackage[utf8]{inputenc}

\title{Towards the Neuroevolution of Low-level Artificial General Intelligence}

\author{Sidney Pontes-Filho$^{a,b,*}$, Kristoffer Olsen$^{c}$, Anis Yazidi$^{a,d,e}$, Michael A. Riegler$^{f}$, Pål Halvorsen$^{a,f}$ and\\
\textbf{Stefano Nichele$^{a,d,e,f,g}$}
\bigskip \\ 
$^a$Department of Computer Science, Oslo Metropolitan University, Oslo, Norway\\
$^b$Department of Computer Science, Norwegian University of Science and Technology, Trondheim, Norway\\
$^c$Department of Informatics, University of Oslo, Oslo, Norway\\
$^d${\it AI Lab} -- OsloMet Artificial Intelligence Lab, Oslo, Norway\\
$^e${\it NordSTAR} -- Nordic Center for Sustainable and Trustworthy AI Research, Oslo, Norway\\
$^f$Department of Holistic Systems, Simula Metropolitan Centre for Digital Engineering, Oslo, Norway\\
$^g$Department of Computer Science and Communication, Østfold University College, Halden, Norway\\
$^*$Corresponding author: sidneyp@oslomet.no}

\date{\small}

\begin{document}
\maketitle
\begin{abstract}
In this work, we argue that the search for Artificial General Intelligence (AGI) should start from a much lower level than human-level intelligence. The circumstances of intelligent behavior in nature resulted from an organism interacting with its surrounding environment, which could change over time and exert pressure on the organism to allow for learning of new behaviors or environment models. Our hypothesis is that learning occurs through interpreting sensory feedback when an agent acts in an environment. For that to happen, a body and a reactive environment are needed. We evaluate a method to evolve a biologically-inspired artificial neural network that learns from environment reactions named Neuroevolution of Artificial General Intelligence (NAGI), a framework for low-level AGI. This method allows the evolutionary complexification of a randomly-initialized spiking neural network with adaptive synapses, which controls agents instantiated in mutable environments. Such a configuration allows us to benchmark the adaptivity and generality of the controllers. The chosen tasks in the mutable environments are food foraging, emulation of logic gates, and cart-pole balancing. The three tasks are successfully solved with rather small network topologies and therefore it opens up the possibility of experimenting with more complex tasks and scenarios where curriculum learning is beneficial.
\end{abstract}

\section{Introduction}

Artificial General Intelligence (AGI) or strong Artificial Intelligence (AI) is commonly discussed among AI researchers. It is often defined as human-level AI, however, the generality of an AI does not need to be considered at such a level of complexity. In fact, our current artificial intelligent systems cannot emulate the learning capabilities of an animal with a simple nervous system, such as a worm \citep{ardiel2010elegant,randi2020measuring}. An alternative approach is to start the quest for the generality of AI from the simplest tasks that animals can do, but machines cannot \citep{crosby2019animal}. Moreover, AGI systems should be tested in tasks that require self-learning on the fly from sensory feedback, as it is often done in meta-learning and continual learning \citep{najarro2020meta,zohora2021metaplasticnet}.

We argue that a radical paradigm change is needed in order to reach general intelligence \citep{crosby2019animal,lake2017building}. Our hypothesis is that such a new paradigm requires learning systems with self-organizing properties, as discussed by \citet{risi2021future}. In this work, our goal is to achieve the learning capabilities of a primitive brain. Therefore, we aim at a low-level AGI, i.e., a system that can learn a map function through sensory experience. Interpreting and understanding sensory inputs are achieved through evolution, particularly supervised evolution \citep{zador2019critique} of agents interacting with their environment.

The brain is the organ that interprets the encoded signals from our sensory organs, thanks to the ability to distinguish between positive and negative sensory experiences depending on what is considered to be good or harmful, e.g., pleasure and pain. Respectively, pleasure and pain serve as reward and penalty mechanisms that affect our behavior.

In this work, we evaluate the Neuroevolution of Artificial General Intelligence (NAGI) framework \citep{pontes2019conceptual}. NAGI is a low-level biologically-inspired AGI framework. NAGI consists of an evolvable spiking neural network with adaptive synapses and randomly-initialized weights. The network is evolved by an extension of the method NeuroEvolution of Augmenting Topologies (NEAT) \citep{stanley2002evolving}. The source code of NAGI is available at \url{https://github.com/SocratesNFR/neat-nagi-python}.

The evolved spiking neural network controls an agent placed in a mutable environment. Its chances of reproduction are proportional to how long it can survive in an environment that is constantly changing, sometimes abruptly. Evolution optimizes how the neurons are connected in the network, their type of neurotransmitters (excitatory or inhibitory), their susceptibility to background electrical current noise (analogous to bias), and their neuroplasticity. With such degrees of freedom in the optimization process, we attempt to approximately recapitulate the evolutionary process of the simplest brains. The mutable environment and random weight initialization propitiate a benchmark for generality and adaptivity of the agent. 

We test NAGI in three mutable environments. The first one is a simple food foraging task, in which the agent has one photoreceptor (or light intensity sensor) used to identify food. The food type (color) is either black or white. Food can be edible or poisonous and this feature changes over time. The agent can also taste the food as its sensory feedback for good and bad actions. The second environment is a logic gate task. The spiking neural network needs to emulate different logic gates in series where the only reward and penalty sensory signals are the supporting mechanisms to identify the correct output. The third environment is a cart-pole balancing task. In this environment, the goal of the agent is to control the forces applied to the cart in order to maintain the pole over it upright. The mutable component of this environment is the pole length which changes during the lifetime of the agent. Because this environment has sensory feedback for the agent's actions, there is no need to add reward and penalty sensory signals.

The article is organized as follows: Section~\ref{sec:background} explains the theoretical basis for understanding NAGI. Section~\ref{sec:related} discusses the related work to our approach. Section~\ref{sec:nagi} describes the details of the method and experiments. Section~\ref{sec:results} presents the experimental results. Section~\ref{sec:discussion} concludes the article including a discussion of the results and plans for future work.

\section{Background}
\label{sec:background}

The components of the NAGI framework are inspired by the overlapping research fields of artificial life \citep{langton2019artificial}, evolutionary robotics \citep{doncieux2015evolutionary}, and computational neuroscience \citep{trappenberg2009fundamentals}. In particular, the controller for the agents is a Spiking Neural Network (SNN) \citep{izhikevich2003simple}, which is a more biologically-plausible artificial neural network. The neurons in an SNN communicate through spikes, i.e., binary values in time series. Therefore, an SNN adds a temporal dimension to binary data. A neuron propagates such data depending on whether its membrane potential crossed a threshold value or not. If the threshold is crossed, the neuron propagates a signal represented as neurotransmitters to its connected neurons; otherwise, the action potential is not propagated. When neurotransmitters are released by a neuron, they can be of two types: excitatory, which increases the membrane potential and the likelihood of producing an action potential; or inhibitory, which has the opposite effect by decreasing the membrane potential. Efficient optimization of an SNN cannot happen through gradient descent as spike trains are not differentiable \citep{tavanaei2019deep}. Instead, spiking neurons have biologically inspired local learning rules, such as Hebbian learning and Spike-Timing-Dependent Plasticity (STDP) \citep{hebb-organization-of-behavior-1949,li2014activity}. Those neuroplasticity rules are unsupervised. Their functionality in the brain is still not fully understood. However, it is inferred that the supervision comes from a certain network configuration acquired through evolution. Therefore, in this work, we use a modification of NeuroEvolution of Augmenting Topologies (NEAT) \citep{stanley2002evolving}. NEAT uses a Genetic Algorithm (GA) \citep{holland1992genetic} to optimize the weights and the topology of a growing neural network that is initialized with a minimal and functional size. NEAT is typically used to search for a network configuration that improves a fitness score while maintaining population diversity (speciation) and avoiding loss of genes during crossover (historical marking). The weights in the NAGI framework are randomly initialized, and they change (adapt) after deployment. The adaptation is coordinated by a realistic Hebbian learning rule, i.e., STDP. This neuroplasticity adjusts the synaptic strength of a neuron's dendrites (i.e., input connections) when it fires an action potential (or spike) that goes through its axon (i.e., output connection). The weights are modified according to the difference in time between incoming spikes and the generated action potential.

The body and brain interaction (sensors and actuators vs. controller) is often described as "chicken and egg" problem \citep{funes1998evolutionary}. The natural evolution of body and brain happens together with the evolution of the environment. They evolve in cooperation and response to each other \citep{mautner2000evolving}. The application of supervised evolution of agents interacting with the environment is defined as embodied evolution \citep{Watson1999EmbodiedEE}. As such, an agent needs a body to learn from the reaction of its environment. We hypothesize that low-level general intelligence in nature emerged through the evolution of a sensory feedback learning method.

\section{Related work}
\label{sec:related}

Neuroevolution with adaptive synapses was introduced in 2003 by \citet{stanley2003evolving}. Such a method is a version of NEAT where the synaptic strength of the connections changes with Hebbian local learning rules. In their work, they used a food foraging task where an agent moves around a field surrounded by edible and poisonous food. The type of food did not change over time, but it was initialized differently at every new run. The agents needed to try the food first before identifying it. Therefore, the agents possess reward and penalty sensory signals as in NAGI. \citet{risi2010indirectly} proposed an extended version by replacing the direct encoding of the network in NEAT with an indirect encoding.


Additional related methods are Refs.~\citep{gaier2019weight} and \citep{najarro2020meta} where  randomly-initialized artificial neural networks are used. The work of \citet{gaier2019weight} uses a version of NEAT where each neuron can have one activation function out of several types. While in the method of \citet{najarro2020meta}, the network topology is fixed and each connection evolves to optimize the parameters of its Hebbian learning rule.

In a recent review on neuroevolution \citep{Stanley2019}, NEAT and its extensions are comparable to deep neural networks trained with gradient-based methods for reinforcement learning tasks. Such methods allow evolving artificial neural networks with indirect encoding for scalability, novelty search for diversity, meta-learning for learning how to learn, and architecture search for deep learning models. Moreover, neuroevolution is described as a key factor for reaching AGI, in particular in relation to meta-learning and open-ended evolution (OOE). Meta-learning encompasses the training of a model with certain datasets and testing with others. The goal of the model is therefore to learn any given dataset by itself from experience \citep{thrun1998learning}. OOE is the ability to endlessly generate a variety of solutions of increasing complexity \citep{taylor2019ooe}. In NAGI, meta-learning is an implicit target in the mutable environments and is implemented as neuroplasticity in the spiking neural network. 

In 2020, \citet{nadji2020brain} introduced the framework for evolutionary artificial general intelligence (FEAGI). This method uses an indirect encoding technique for a spiking neural network that resembles the growth of the biological brain, which is called "neuroembryogenesis". As a proof of concept, FEAGI demonstrates successful handwritten digits classification by learning through association and being able to recall digits from different image samples in real time.


\section{Neuroevolution of Artificial General Intelligence}
\label{sec:nagi}

The NAGI framework aims at providing a simplified model of the initial stages of the evolution of biological general intelligence \citep{pontes2019conceptual}. The evolving agents in NAGI consist of randomly-initialized spiking neural networks. Thus, a genome in NAGI does not require the definition of synaptic weights of the connections between neurons, as it is done in NEAT. Therefore, the synaptic weights in the genome are replaced by an STDP rule and its parameters for each neuron. Since biological neurons may provide one of the two main neurotransmitters, NAGI's genome defines such a feature in the neurons' genes. As such, a neuron can be either excitatory or inhibitory. To imitate the function of bias in artificial neural networks, neurons may be also susceptible to a "background electrical current noise".

The environment changes during the lifetime of the agent. This forces the agent to learn new environmental conditions. Therefore, the agent is encouraged to generalize and learn how to learn. The aforementioned random initialization and mutable environment aim at benchmarking the basic properties needed for low-level AGI.

\subsection{Spiking neural network}

The spiking neural network has a fixed number of input and output neurons depending on the task to be solved. The neuroevolution process defines the number of hidden neurons that will be available. Hidden neurons can be either excitatory or inhibitory, while input and output neurons are always excitatory. Self-loops and cycles are permitted while duplicate connections between two neurons in the same direction are prohibited. The SNN is stimulated from the input neurons, as such units are spike generators. The spikes are uniformly generated in an assigned frequency or firing rate.

As a spiking neuron model, we use a simplification of the leaky integrate-and-fire model \citep{liu2001spike}. A neuron's membrane potential $v$ is increased directly by its inputs and decays over time by a factor $\lambda_{decay}$. We can then express the change in membrane potential $\Delta v$ with regards to a time step $\Delta t$ by
\begin{equation}
\label{eq:lif}
    \Delta v(\Delta t) = \sum_{i=1}^{n}w_i x_i - \Delta t \lambda_{decay} v,
\end{equation}
where $x_i$ is the input value $0$ (no spike) or $1$ (spike) from the presynaptic neuron $i$, the dendrite for this connection has the synaptic strength defined as $w_i$, and $n$ is the total number of presynaptic neurons that the dendrites are connecting. If the membrane potential $v$ is greater than the membrane threshold $v_{th}$, a spike is released and the membrane potential returns to the resting membrane potential $v_{rest}$, which is $0$. The time step $\Delta t$ we use in the experiments is $0.1$ ms, and decay factor $\lambda_{decay}$ is $0.01\Delta t$. An action performed by the SNN is calculated by the number of spikes in a time window. Such an actuator time window covers 250 ms or 2,500 time steps. In NAGI, the weights of the SNN are randomly initialized with a normal distribution. The mean is equal to $1$ and the standard deviation is equal to $0.2$. The weights are always positive. As mentioned, the excitation and inhibition of a neuron are defined by the neurotransmitter of the presynaptic neuron.

\subsubsection{Homeostasis}

Biological neurons have a plasticity mechanism that maintains a steady equilibrium of the firing rate, which is called homeostasis \citep{betts2013anatomy,kulik2019dual}. In our method, the spiking neurons can have non-homogeneous inputs, which could lead to very different firing rates. It is desirable that all neurons have approximately equal firing rates \citep{homeostasis}. In order to homogenize the firing rates of the neurons in a network, the membrane threshold $v_{th}^{*}$ is given by 
\begin{equation}
    v_{th} = \min(v_{th}^{*} + \Theta,\sum_{i=1}^{n}w_i),
\end{equation}
where $v_{th}^{*}$ is the "resting" membrane threshold equals to $1$, and $\Theta$ starts with value $0$, is increased every time a neuron fires by a value of $0.2$ and decays exponentially with a rate of $0.01\Delta t$. Each neuron has an individual $\Theta$. Therefore, a neuron firing more often will get a larger membrane threshold and as consequence a lower firing rate. To compensate for a neuron with weak incoming weights, which causes a low firing rate; we use the sum of the incoming weights as the threshold instead.

\subsubsection{Spike-Timing-Dependent Plasticity}

The adjustment of the weights of the connections entering into a neuron happens on every input and output spike to and from a neuron. This is performed by STDP and is done by keeping track of the time elapsed since the last output spike, as well as the time elapsed since each input spike for each incoming connection within a time frame, called the STDP time window, which is set to be $\pm40$ ms. The difference between presynaptic and postsynaptic spikes, or the relative timing between them, denoted by $\Delta t_{r}$ is given by
\begin{equation}
    \Delta t_{r}(t_{out}, t_{in}) = t_{out} - t_{in},
\end{equation}
where $t_{out}$ is the timing of the output spike and $t_{in}$ is the timing of the input spike.

The synaptic weight change $\Delta w$ is calculated in accordance with one of the four Hebbian learning rules. The functions for each of the four learning rules are given by
\begin{align}
    \Delta w(\Delta t_{r}) &= \left\{
        \begin{array}{ll}
            A_{+}e^{\frac{-\Delta t_{r}}{\tau_{+}}} & \Delta t_{r} > 0, \\
            -A_{-}e^{\frac{\Delta t_{r}}{\tau_{-}}} & \Delta t_{r} < 0, \\
            0 & \Delta t_{r} = 0; \\
        \end{array} 
    \right. & \text{Asymmetric Hebbian} \\
    \Delta w(\Delta t_{r}) &= \left\{
        \begin{array}{ll}
            -A_{+}e^{\frac{-\Delta t_{r}}{\tau_{+}}} & \Delta t_{r} > 0, \\
            A_{-}e^{\frac{\Delta t_{r}}{\tau_{-}}} & \Delta t_{r} < 0, \\
            0 & \Delta t_{r} = 0; \\
        \end{array} 
    \right. & \text{Asymmetric Anti-Hebbian} \\
    \Delta w(\Delta t_{r}) &= \left\{
        \begin{array}{ll}
            A_{+}g(\Delta t_{r}) & g(\Delta t_{r}) > 0,\\
            A_{-}g(\Delta t_{r}) & g(\Delta t_{r}) < 0, \\
            0 & g(\Delta t_{r}) = 0; \\
        \end{array} 
    \right. & \text{Symmetric Hebbian} \\
    \Delta w(\Delta t_{r}) &= \left\{
        \begin{array}{ll}
            -A_{+}g(\Delta t_{r}) & g(\Delta t_{r}) > 0, \\
            -A_{-}g(\Delta t_{r}) & g(\Delta t_{r}) < 0, \\
            0 & g(\Delta t_{r}) = 0; \\
        \end{array}
    \right. & \text{Symmetric Anti-Hebbian}
\end{align} 
where $g(\Delta t_{r})$ is a Difference of Gaussian function given by
\begin{equation}
    g(\Delta t_{r}) = \frac{1}{\sigma_{+}\sqrt{2\pi}}e^{-\frac{1}{2}(\frac{\Delta t_{r}}{\sigma_{+}})^{2}} - \frac{1}{\sigma_{-}\sqrt{2\pi}}e^{-\frac{1}{2}(\frac{\Delta t_{r}}{\sigma_{-}})^{2}},
\end{equation}

\begin{figure}[!htb]
\centering
  \begin{subfigure}{0.24\textwidth}
    \includegraphics[width=\linewidth]{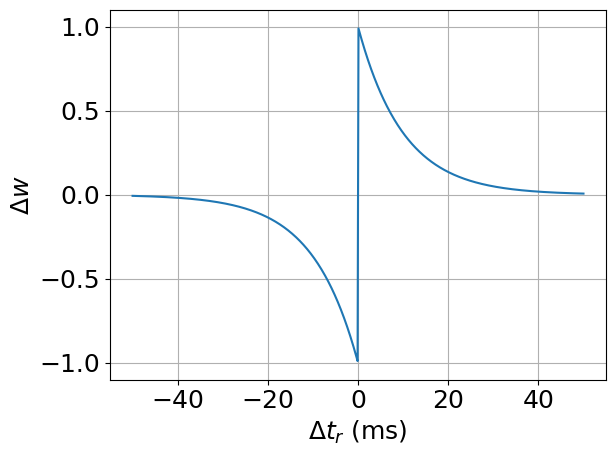}
    \caption{Asymmetric Hebbian} \label{fig:asymmetric_hebbian}
  \end{subfigure} \hspace*{\fill} 
  \begin{subfigure}{0.24\textwidth}
    \includegraphics[width=\linewidth]{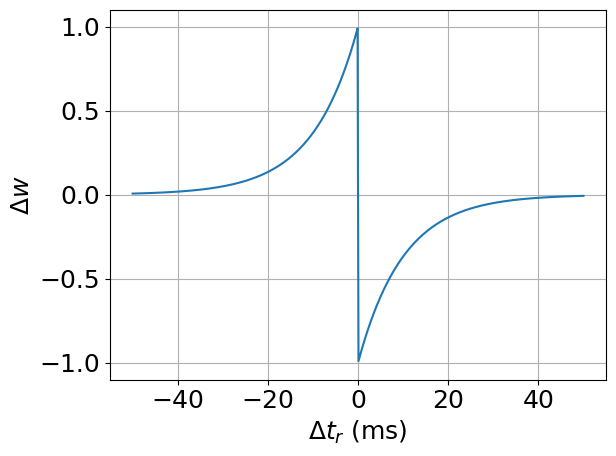}
    \caption{Asymmetric Anti-Hebbian} \label{fig:asymmetric_anti_hebbian}
  \end{subfigure} \hspace*{\fill} 
  \begin{subfigure}{0.24\textwidth}
    \includegraphics[width=\linewidth]{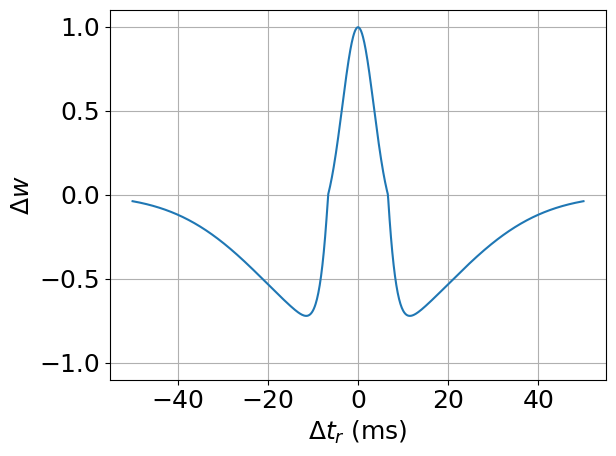}
    \caption{Symmetric Hebbian} \label{fig:symmetric_hebbian}
  \end{subfigure} \hspace*{\fill} 
  \begin{subfigure}{0.24\textwidth}
    \includegraphics[width=\linewidth]{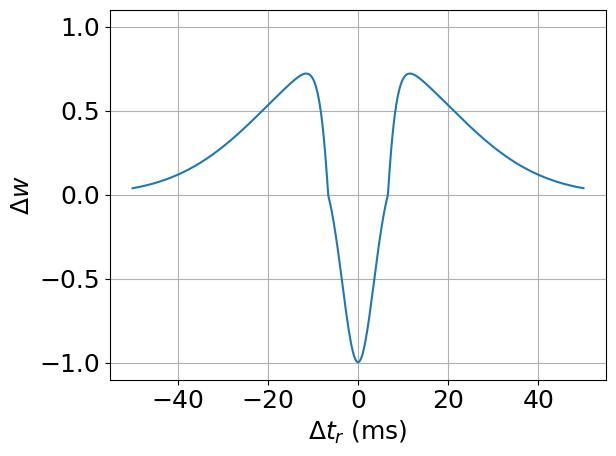}
    \caption{Symmetric Anti-Hebbian} \label{fig:symmetric_anti_hebbian}
  \end{subfigure} 
\caption{\label{fig:stdp} Spike-timing-dependent plasticity rules.}
\end{figure}

$A_{+}$ and $A_{-}$ are the parameters that affect the height of the curve, $\tau_{+}$ and $\tau_{+}$ are the parameters that affect the width or steepness of the curve of the Asymmetric Hebbian functions, and $\sigma_{+}$ and $\sigma{-}$ are the standard deviations for the Gaussian functions used in the Symmetric Hebbian functions. It is also required that $\sigma_{-} > \sigma_{+}$. We experimentally found fitting ranges for each of these parameters, which can be seen in Tab.~\ref{tab:asymmetric} and Tab.~\ref{tab:symmetric}. The STDP curves with the maximum value of those parameters are illustrated in Fig.~\ref{fig:stdp}.
\begin{table}[ht]
    \centering
    \begin{tabular}{ |c|c| }
        \hline
        \multicolumn{2}{|c|}{Asymmetric} \\
        \hline
        $A_{+}$ & $[0.1, 1.0]$ \\  
        \hline
        $A_{-}$ & $[0.1, 1.0]$ \\
        \hline
        $\tau_{+}$ & $[1.0, 10.0]$ \\
        \hline
        $\tau_{-}$ & $[1.0, 10.0]$\\
        \hline
    \end{tabular}
    \caption{\label{tab:asymmetric} Asymmetric STDP parameter ranges.}
\end{table}
\begin{table}[ht]
    \centering
    \begin{tabular}{ |c|c| }
        \hline
        \multicolumn{2}{|c|}{Symmetric} \\
        \hline
        $A_{+}$ & $[1.0, 10.6]$ \\  
        \hline
        $A_{-}$ & $[1.0, 44.0]$ \\
        \hline
        $\sigma_{+}$ & $[3.5, 10.0]$ \\
        \hline
        $\sigma_{-}$ & $[13.5, 20.0]$\\
        \hline
    \end{tabular}
    \caption{\label{tab:symmetric} Symmetric STDP parameter ranges.}
\end{table}

Weights can take values in a range $[w_{min}, w_{max}]$, and every neuron has a weight budget $w_{budget}$ it must follow. What this means is that if the sum of a neuron's incoming weights exceed $w_{budget}$ after initialization or STDP has been applied, they are normalized to $w_{budget}$, given by 
\begin{equation}
    \textrm{if $\sum_{i=1}^{n}w_i > w_{budget}$, then } w_{i} = \frac{w_{i}w_{budget}}{\sum_{i=1}^{n}w_i}.
\end{equation}
The parameters used during our experiments are $w_{min}=0$, $w_{max}=1$, and $w_{budget}=5$. In the SNN without homeostasis, if a connection $i$ has $w_i=w_{max}$, then $w_i=v_{th}$. Therefore, an action potential coming from $i$ will always produce a spike. This is the reason to have $w_{max}=v_{th}$.

\subsection{Genome}

The genome in NAGI is rather similar to the one in NEAT. Its node genes have three types: input, hidden, and output. Depending on the type of the node gene, there is a different collection of \textit{loci}\footnote{In the terminology of genetic algorithms, a value within a gene is also called a \textit{locus} (plural \textit{loci}).}. The input node is a spike generator and provides excitation to the neurons it is connected to. The gene of an input node is the same as in NEAT. The hidden and output nodes represent adaptable and mutable spiking neurons. They have three additional loci: the type of the learning rule, the set of the learning rule parameters, and a bias. The connection gene in NAGI has no weight locus as in NEAT. The reason for its removal is that the weights of the SNN are defined by a normal distribution.

The learning rule is one of the four STDPs. The set of learning rule parameters consists of four parameters that adjust the intensity of the weight change. They are different for symmetric and asymmetric learning rules. The symmetric parameters are $\{A_{+}$, $A_{-}$, $\sigma_{+}, \sigma_{-}\}$ and the asymmetric parameters are $\{A_{+}$, $A_{-}$, $\tau_{+}, \tau_{-}\}$. The bias is a Boolean value that determines if the neuron has a constant input of $0.001$ being added to $\Delta v$, which is analogous to the background noise of the neuron.

\begin{figure*}[htb]
\includegraphics[width=0.7\textwidth]{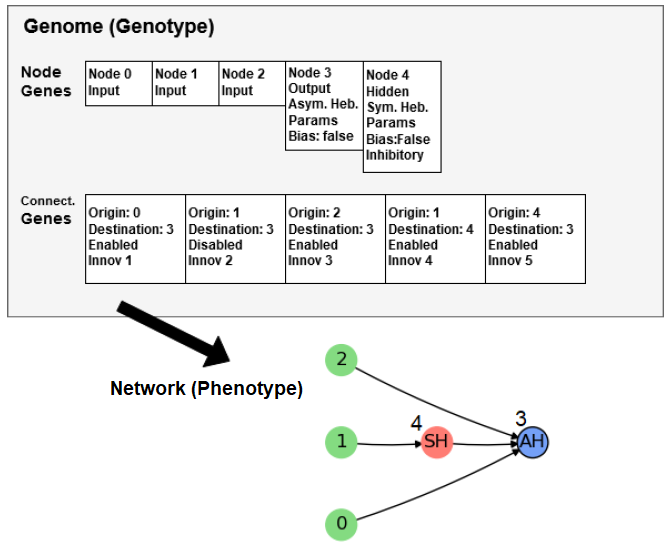}
\centering
\caption{Genotype and phenotype in NAGI. Image taken from Ref.~\citep{olsen2020neuroevolution}.}
\label{fig:genotype}
\end{figure*}

The hidden node genes have a unique locus, which is a Boolean value that determines whether it represents an inhibitory or excitatory neuron. This locus is not included in the output node genes because they are always excitatory. As a result of combining all the descriptions of the genome in NAGI, the genotype and the phenotype are illustrated in Fig.~\ref{fig:genotype}.

The initialization of the additional loci in the node genes can be conditional and non-uniform. The initialization of the neurotransmitter type of a neuron follows a similar proportion of excitatory and inhibitory neurons in the brain \citep{sukenik2021neuronal}. The probability of a neuron being added as excitatory is 70\%. The probability of having a bias is 20\%. Depending on the neurotransmitter, excitatory neurons have a 70\% chance of initializing with Hebbian plasticity, and inhibitory neurons have the same chance but for anti-Hebbian plasticity. The learning rule parameters are initialized by sampling from a uniform distribution within the ranges in Tabs.~\ref{tab:asymmetric} and \ref{tab:symmetric}.

The mutations of the additional loci happen in 10\% of chance to switch the neurotransmitter type, bias, learning rule, and learning rule parameters. Those parameters have 2\% chance of a fully re-initialization. When the parameters are assigned to be mutated, a random value sampled from a normal distribution with $\mu = 0$ and $\sigma^{2}=m(p)$ is added to the parameter $p$. The equation of $m(p)$ is 
\begin{equation}
m(p) = 0.2(p_{max} - p_{min}),
\end{equation}
where $p_{max}$ and $p_{min}$ are the maximum and minimum values the parameter can have, given by the ranges in Tabs.~\ref{tab:asymmetric} and \ref{tab:symmetric}. During the neuroevolution, 10\% of the genotypes with the best fitness scores will be passed to the next generation unchanged, i.e., elitism.

\subsection{Mutable environments}

The benchmark tasks for NAGI are meant to evaluate the agent's ability to generalize and self-adapt. Therefore, they consist of environments that change during the lifetime of the agent. Two types of tasks are provided, binary classification (two tasks of this kind are provided) and control (one task of this kind is provided). The first type (binary classification) is the simplest one, however, it provides the most abrupt changes in the environment. The binary classification tasks are food foraging with one input, and logic gates with two inputs. The control task in a simulated physical environment is the cart-pole balancing from OpenAI Gym \citep{openaigym}. The changes are less abrupt in this last task as they consist in modifying the pole size. The fitness scores are calculated using the number of time steps $t$ that the agent survived in these environments, normalized to the range $[0,1]$ using the maximum possible lifetime $L_{max}$ and minimum possible lifetime $L_{min}$. Therefore, the fitness function $f$ is given by
\begin{equation}
\label{eq:fitness}
    f(t) = \frac{t - L_{min}}{L_{max} - L_{min}}.
\end{equation}

In the binary classification tasks, the agents have an initial amount of health points that is reduced every time step as continuous damage. If a correct action is chosen, the health point amount is reduced by $d_c$ health point. Otherwise, it is reduced by $d_i$. The input sample is given to the agent for 1 second or 10,000 time steps, then it is changed to a new one. The mutation of the environment condition happens when the agent has seen four samples. The order of the input samples and the environment conditions is fixed and cyclic.

We noticed that the number of spikes within the actuator time window can be the same for the output neurons and therefore allowing for a tie in many cases. Our solution to avoid spiking neural networks with this behavior is to include a "confidence" factor in the fitness score calculation. Therefore, the higher the difference between the spike count, the more confident the action is. If the action is correct and highly confident, the damage is $d_c$ or closer. If the action is incorrect but highly confident, the damage is $d_i$ or closer. The lack of confidence would make the damage lie between the values $d_c$ and $d_i$. The spike count for the correct action $s_c$ and incorrect one $s_i$ are used to calculate the participation of the spikes for deciding the correct action $p_c$ and the participation for the incorrect action $p_i$. In the iterations without spikes of the output neurons, normally the initial ones; the agent takes $d_i$ as damage. Otherwise, the damage is calculated by
\begin{align}
    p_{c}(s_{c}, s_{i}) &= \left\{
    \begin{array}{ll}
        \frac{\max(0, \min(s_{c}, s_{t})) - \max(0, \min(s_{c}, s_{t})) + s_{t}}{2s_{t}} & s_{c} + s_{i} \leq 2s_{t} \\
        \frac{s_{c}}{s_{c} + s_{i}} & s_{c} + s_{i} > 2s_{t} \\
    \end{array} 
    \right. \\
    p_{i}(s_{c}, s_{i}) &= 1 - p_c(s_{c} - p_{i}) 
\end{align}
where $s_{t}$ is the minimum `target' number of spikes. The purpose of $s_{t}$ is to avoid assigning a too high or low fitness to agents that fire few spikes through their outputs. The agent takes damage at every time step and is given by 
\begin{equation}
    d(s_{c}, s_{i}) = d_{c}p_{c}(s_{c}, s_{i}) + d_{i}p_{i}(s_{c}, s_{i})
\end{equation}
Damaging is performed until the agent runs out of health points and `dies'. Subsequently, the fitness score of the agent is calculated from the fitness function expressed in \eqref{eq:fitness}. The damage to the health points in a correct action $d_c$ is 1, in an incorrect one $d_i$ is 2. Therefore, correct actions result in a longer lifetime. The value for the minimum `target' number of spikes $s_t$ is 3 spikes.

In the control task of cart-pole balancing, the behavior of the mutable environment is different. A new environment is presented to the agent either after its failure or after the maximum number of environment iterations is reached. Moreover, the agents do not have health points. The fitness score is the normalization of the number of iterations that the agent survived after all environment conditions were executed.

\subsubsection{Food foraging}

\begin{figure*}[htb]
\includegraphics[width=0.7\textwidth]{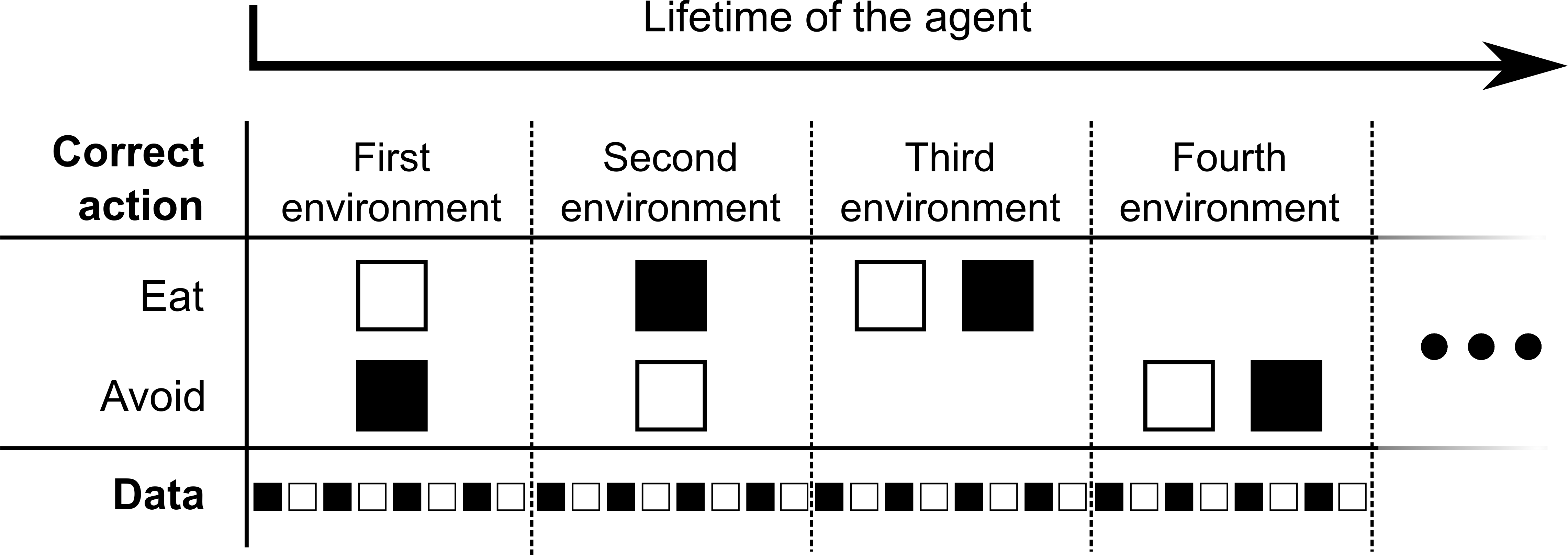}
\centering
\caption{Example of the food foraging environment and how it progresses through the lifetime of the agent in a generation.}
\label{fig:1d_env}
\end{figure*}

The agent in the food foraging environment possesses just one light sensor for identifying the food "in front of it". There are two types of food: edible and poisonous. As such, food is represented in two colors: black and white. The environment changes by randomly defining which food color is edible or poisonous. In this environment, the agent can act in two ways: eating or avoiding the food. The sample has a predefined time of exposure to the agent. An action is performed after the first spike and it continues for every time step in the environment simulation. After this exposure time, the food is replaced by a new one. The agent can only discover whether it is exposed to an edible or poisonous food by interacting with it. An incorrect action is defined as eating poisonous food, or avoiding edible food, while a correct action is defined as eating edible food or avoiding poisonous one. If the agent makes an incorrect action, it receives a penalty signal, from which the agent should learn over the generations that it represents pain, revulsion, or hunger. If the agent makes a correct action, it receives a reward signal, from which it should learn that it represents the pleasure of eating delicious food or recognizing that the food is poisonous. In Fig.~\ref{fig:1d_env}, the food foraging environment is illustrated, how the environment changes and provides new food samples. In Tab.~\ref{tab:food}, the four combinations of edible and poisonous food for the white and black ones are shown. To evolve the spiking neural network for the food foraging task, the parameters of the genetic algorithm are the following: the population size is set to 100 individuals, and the number of generations is set to 1,000.

\begin{table}[htb]
    \centering
    \begin{tabular}{ |c|c c c c| }
    \hline
    \multicolumn{5}{|c|}{Food Foraging Environment Conditions} \\
    \hline
    \diagbox{Input}{Edible} & Black & White & None & Both \\ 
    \hline
    Black & Eat & Avoid & Avoid & Eat \\
    White & Avoid & Eat & Avoid & Eat \\
    \hline
    \end{tabular}
    \caption{\label{tab:food} Correct actions for all combinations of input food color and edible food in the food foraging task.}
\end{table}

\subsubsection{Logic gates}

In this environment, the mutable environmental state is a two-input logic gate. The environment provides the agent with two binary inputs, i.e., 0's and 1's. The agent's task is to predict the correct output for the current logic gate given the current input. Similar to the food foraging environment, it receives a reward signal if it is currently predicting the correct output, and a penalty signal if it is currently predicting the wrong output.

In order to measure the generalizing properties of agents, we use two different sets of environments: a training environment, which is used in calculating the fitness score while running the evolutionary algorithm, and a test environment which has a fully disjoint set of possible environmental states. A full overview of the logic gates found in both the training and the test environments, as well as the truth values for all input and output combinations, are found in Tabs.~\ref{tab:training-gates} and \ref{tab:testing-gates}. The evolution of the spiking neural network is performed by a population of 100 individuals through 1,000 generations.
\begin{table}[ht]
    \centering
    \begin{tabular}{ |c c|c c c c c c c c| }
    \hline
    \multicolumn{10}{|c|}{Training Logic Gates} \\
    \hline
    \multicolumn{2}{|c|}{Input} & \multirow{2}{*}{A} & \multirow{2}{*}{B} & \multirow{2}{*}{NOT A} & \multirow{2}{*}{NOT B} & \multirow{2}{*}{ONLY 0} & \multirow{2}{*}{ONLY 1} & \multirow{2}{*}{XOR} & \multirow{2}{*}{XNOR} \\
    \cline{1-2}
    A & B & & & & & & & &\\
    \hline
    0 & 0 & 0 & 0 & 1 & 1 & 0 & 1 & 0 & 1 \\
    0 & 1 & 0 & 1 & 1 & 0 & 0 & 1 & 1 & 0 \\
    1 & 0 & 1 & 0 & 0 & 1 & 0 & 1 & 1 & 0 \\
    1 & 1 & 1 & 1 & 0 & 0 & 0 & 1 & 0 & 1 \\
    \hline
    \end{tabular}
    \caption{\label{tab:training-gates} Truth table showing the correct output for each training logic gate.}
\end{table}
\begin{table}[ht]
    \centering
    \begin{tabular}{ |c c|c c c c| }
    \hline
    \multicolumn{6}{|c|}{Test Logic Gates} \\
    \hline
    \multicolumn{2}{|c|}{Input} & \multirow{2}{*}{AND} & \multirow{2}{*}{NAND} & \multirow{2}{*}{OR} & \multirow{2}{*}{NOR} \\
    \cline{1-2}
    A & B & & & & \\
    \hline
    0 & 0 & 0 & 1 & 0 & 1\\
    0 & 1 & 0 & 1 & 1 & 0\\
    1 & 0 & 0 & 1 & 1 & 0\\
    1 & 1 & 1 & 0 & 1 & 0\\
    \hline
    \end{tabular}
    \caption{\label{tab:testing-gates} Truth table showing the correct output for each testing logic gate.}
\end{table}

\subsubsection{Cart-pole balancing}

The cart-pole balancing is a well-known control task used as a benchmark problem in reinforcement learning. In this environment, there is a cart that moves when a force is applied to the left or to the right every time step. In the middle of the cart, there is a vertical pole connected to a non-actuated joint. The goal of this environment is to maintain the pole balanced upright by controlling the forces that move the cart. Moreover, the cart cannot move beyond the limits of the track. The observations available to the controller are the cart position, the cart velocity, the pole angle, and the pole angular velocity. 

\begin{figure}[tb]
\centering
\subfloat[Size=0.3]{\label{fig:pole1}\includegraphics[width=0.18\textwidth]{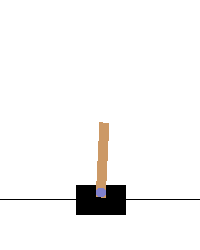}}
\hfill
\subfloat[Size=0.4]{\label{fig:pole2}\includegraphics[width=0.18\textwidth]{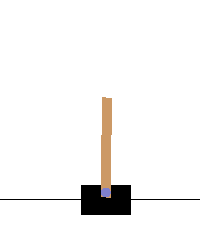}}
\hfill
\subfloat[Size=0.5]{\label{fig:pole3}\includegraphics[width=0.18\textwidth]{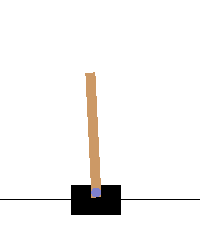}}
\hfill
\subfloat[Size=0.6]{\label{fig:pole4}\includegraphics[width=0.18\textwidth]{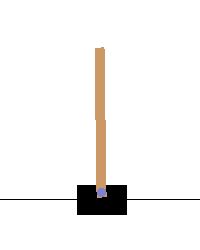}}
\hfill
\subfloat[Size=0.7]{\label{fig:pole5}\includegraphics[width=0.18\textwidth]{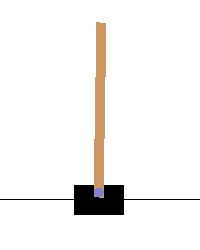}}
\caption{The different pole sizes for the cart-pole balancing task.}
\label{fig:pole}
\end{figure}

For training, we use poles of different sizes, which are $0.5$ (default), $0.3$, and $0.7$. For testing, the sizes are $0.4$, and $0.6$. Those pole sizes are depicted in Fig.~\ref{fig:pole}. Each size can run up to 200 environment iterations and it is repeated three times during training for promoting stable controllers. If there are no more environment iterations or the pole falls, the cart-pole environment restarts with the next pole size while using the same SNN or finishes when all pole sizes were executed. The fitness score is calculated using the number of iterations the pole kept balanced. Subsequently, it is normalized to values between $0$ and $1$. The evolution for this task occurs with a population size of 256 during 500 generations.

\subsection{Data representation}

The data type in a spiking neural network is a binary time series or a spike train. Because the agent senses and acts in the environment, such data must be converted from the sensors and to the actuators. The flow of spikes over time can be quantified as firing rate, which corresponds to a frequency, or the number of spikes per second. The firing rate is the data representation that is converted as inputs and outputs for the SNN. However, the input firing rate must be within a minimum and a maximum value. In our experiments, we use the value range $[5Hz, 50Hz]$. The minimum and maximum value of the firing rate are simplified to a real number range $[0,1]$. It is preferable that the data from the sensors has also a minimum and a maximum value. Otherwise, it will be necessary to clip sensory values or map the values to a desirable range.

In the binary classification tasks, all inputs and outputs are binary. Therefore, the minimum and maximum values for the input firing rate stand for, respectively, $0$ and $1$, or $False$ and $True$. To avoid having a predefined threshold firing rate for the output neurons, we opt to have two output neurons for one binary value. The neuron with the highest firing rate within the actuator time window is the one defining the binary output value. If these two output neurons have the same firing rate, then the last one with the highest value is selected. We also decided to have the same "two neurons-one binary value" strategy with the inputs. Therefore, the inputs are presented as one-hot encoding, as shown in Tab.~\ref{tab:onehot}.

\begin{table}[ht]
    \centering
    \begin{tabular}{ |c|c|c| }
        \hline
        Binary & One-Hot & Firing Rate\\  
        \hline
        0 & 01 & (low, high)\\
        \hline
        1 & 10 & (high, low)\\
        \hline
    \end{tabular}
    \caption{\label{tab:onehot} How binary encoded data values coincide with one-hot encoded values, which in turn translates into a tuple of firing rates for SNN input.}
\end{table}

\begin{figure}[tb]
\centering
\subfloat[Cart position, cart velocity, and pole angular velocity]{\label{fig:sense1}\includegraphics[width=0.45\textwidth]{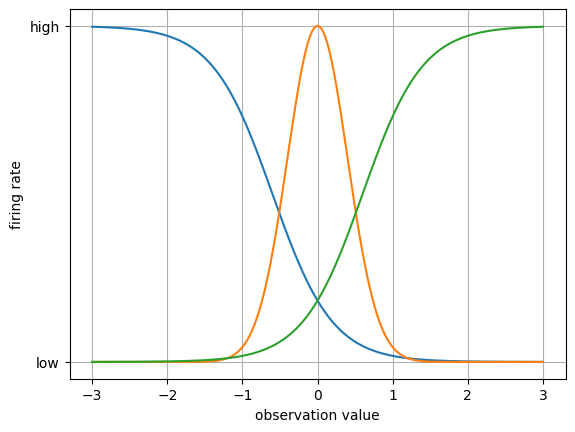}}
\hspace{1cm}
\subfloat[Pole angle (in radians)]{\label{fig:sense2}\includegraphics[width=0.45\textwidth]{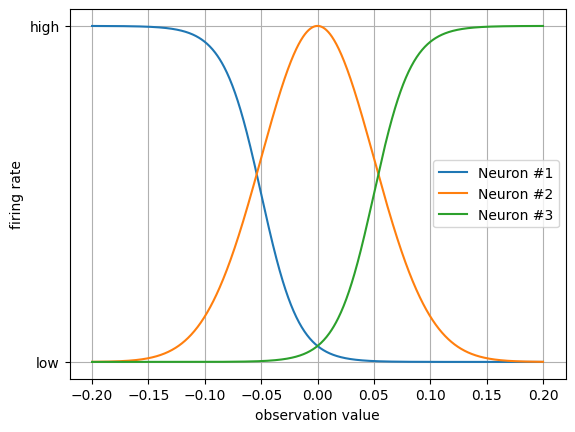}}
\caption{Converting observation values to firing rate of the three input neurons.}
\label{fig:sense}
\end{figure}

For the cart-pole control task, the inputs are real numbers, and the left and right actions are represented as two output neurons, similar to the outputs of the binary classification tasks. In this environment, the inputs are the cart position, cart velocity, pole angle, and pole angular velocity. Because we infer that real numbers converted to the firing rate of one neuron can be difficult to deal with in an adaptive spiking neural network (as also mentioned in Ref.~\citep{pontes2019bidirectional}), we decided to have three neurons for each input. The firing rate of the three neurons is similar to the sensitivity for the light spectrum of the three cone cells in the human eye \citep{Bowmaker1980}. The conversion from observation values to firing rate is depicted in Fig.~\ref{fig:sense}. We use the sigmoid function \citep{sigmoid} for neurons \#1 and \#3 and the a normalized version of the Gaussian function \citep{patel1996gaussian} for neuron \#2. The sigmoid equation is 
\begin{equation}
\label{eq:sigmoid}
{\cal F}_{sigmoid}(x|\omega,z,h,l)=\frac{h}{1+e^{-\omega(x-z)}}+l,
\end{equation}
where $x$ is the observation value from the environment, $\omega$ is the weight that adjusts the smoothness of the interval between $0$ and $1$, $z$ is the shift coefficient to adjust the function on the horizontal axis, $h$ is the highest firing rate possible applied to an input neuron, and $l$ is the lowest firing rate possible. The Gaussian function for converting observation value to firing rate is expressed by
\begin{equation}
\label{eq:gauss}
{\cal F}_{Gaussian}(x|\mu,\sigma,h,l)=he^\frac{-((x-\mu)^2)}{2\sigma ^2}+l,
\end{equation}
where $\mu$ is the mean and $\sigma$ is the standard deviation. We replace $\frac{1}{\sigma\sqrt{2\pi}}$ in the original Gaussian function to $h$ because, in this way, we can define the highest firing rate when the observation value is the mean.

Neurons \#1 and \#3 use ${\cal F}_{sigmoid}$, while neuron \#2 uses ${\cal F}_{Gaussian}$. For cart position, cart velocity, and pole angular velocity, the parameters are shown in Tab.~\ref{tab:cartpole_param1}. This is depicted in Fig.~\ref{fig:sense1}. For pole angle, the parameters are presented in Tab.~\ref{tab:cartpole_param2}. These three functions are depicted in Fig.~\ref{fig:sense2}.

\begin{table}[]
\centering
\begin{tabular}{|l|l|l|}
\hline
Neuron          & Function          & Parameters            \\ \hline
\#1             & ${\cal F}_{sigmoid}$     & $w=-2.5, z=-0.6$      \\ \hline
\#2             & ${\cal F}_{Gaussian}$    & $\mu=0.0, \sigma=0.4$ \\ \hline
\#3             & ${\cal F}_{sigmoid}$     & $w=2.5, z=0.6$        \\ \hline
\end{tabular}
\caption{\label{tab:cartpole_param1} Parameters of ${\cal F}_{sigmoid}$ and ${\cal F}_{Gaussian}$ for cart position, cart velocity, and pole angular velocity.}
\end{table}

\begin{table}[]
\centering
\begin{tabular}{|l|l|l|}
\hline
Neuron          & Function          & Parameters            \\ \hline
\#1             & ${\cal F}_{sigmoid}$     & $w=-60.0, z=-0.05$      \\ \hline
\#2             & ${\cal F}_{Gaussian}$    & $\mu=0.0, \sigma=0.05$ \\ \hline
\#3             & ${\cal F}_{sigmoid}$     & $w=60.0, z=0.05$        \\ \hline
\end{tabular}
\caption{\label{tab:cartpole_param2} Parameters of ${\cal F}_{sigmoid}$ and ${\cal F}_{Gaussian}$ for pole angle.}
\end{table}

\section{Results}
\label{sec:results}

The evolution of the spiking neural networks in NAGI is evaluated with fitness score, accuracy, and end-of-sample accuracy for the binary classification tasks, which are food foraging and logic gate. The accuracy is measured at every time step of the simulation. The end-of-sample accuracy stands for the accuracy measured in the last time step of a sample. The assessment performed for the control task with cart-pole balancing is done with the fitness score. We test the best performing agent in a task with ten simulations where their details are also provided.

\begin{figure}[t]
\centering
\subfloat[Fitness]{\label{fig:food_fit}\includegraphics[width=0.31\textwidth]{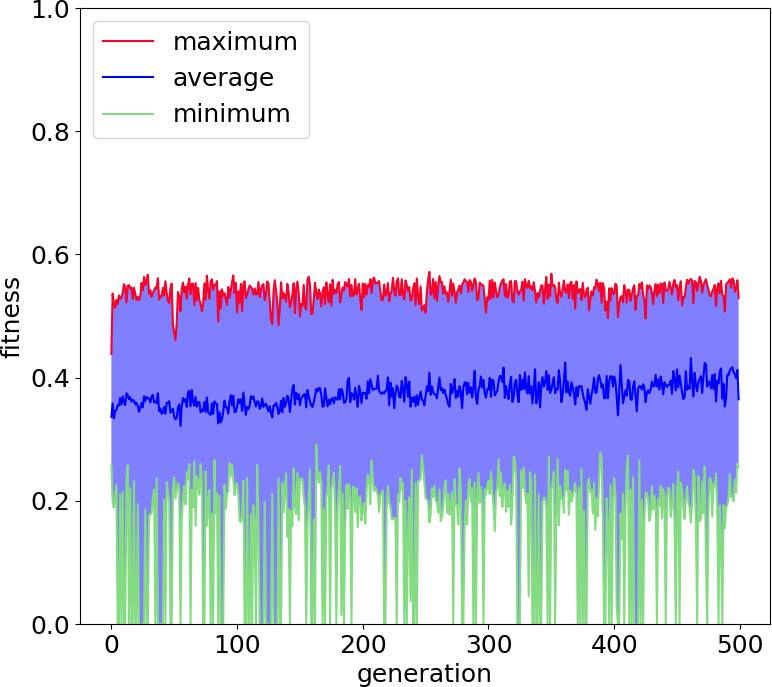}}
\hfill
\subfloat[Accuracy]{\label{fig:food_acc}\includegraphics[width=0.31\textwidth]{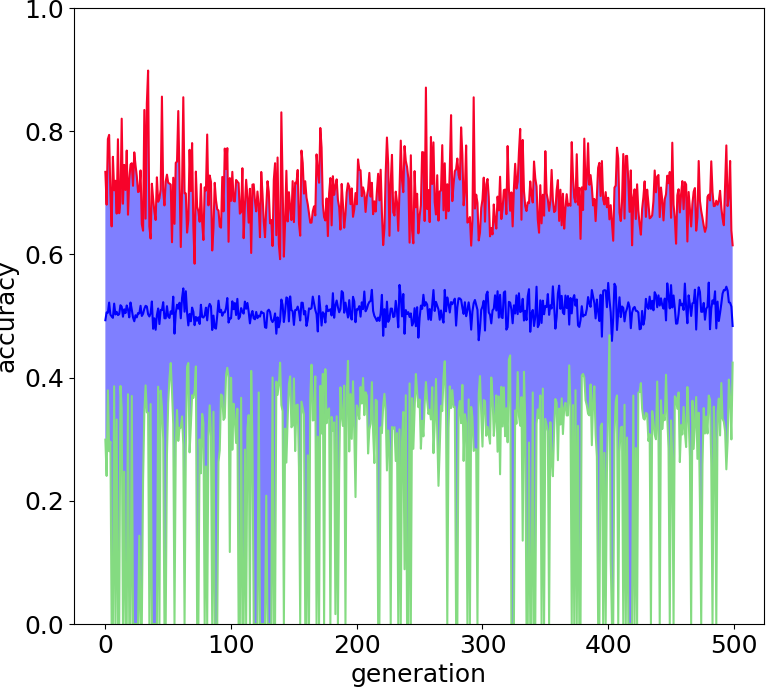}}
\hfill
\subfloat[End-of-sample accuracy]{\label{fig:food_eos_acc}\includegraphics[width=0.31\textwidth]{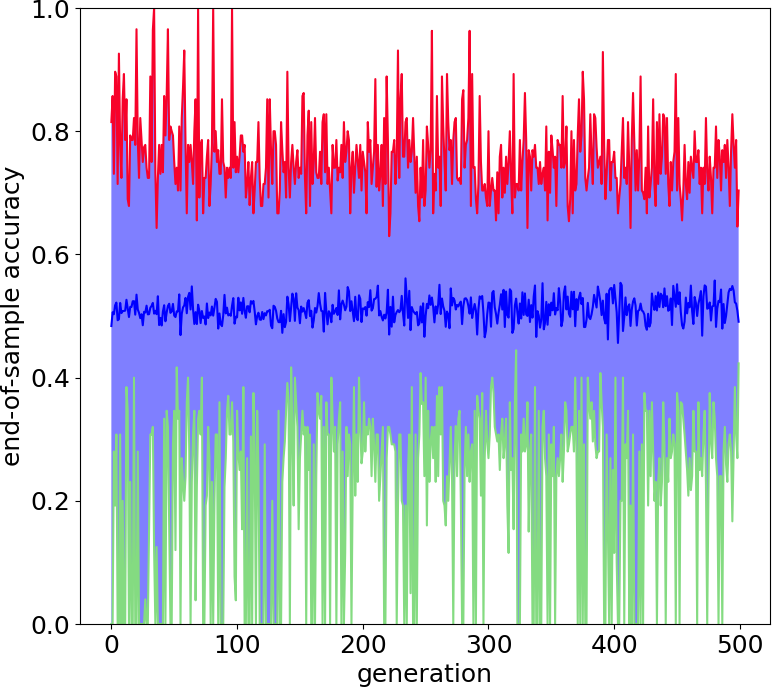}}
\caption{Evolution history of food foraging environment showing the average, minimum and maximum per generation.}
\label{fig:food_res}
\end{figure}

\begin{figure}[!htb]
\centering
  \begin{subfigure}{0.49\textwidth}
    \includegraphics[width=\linewidth]{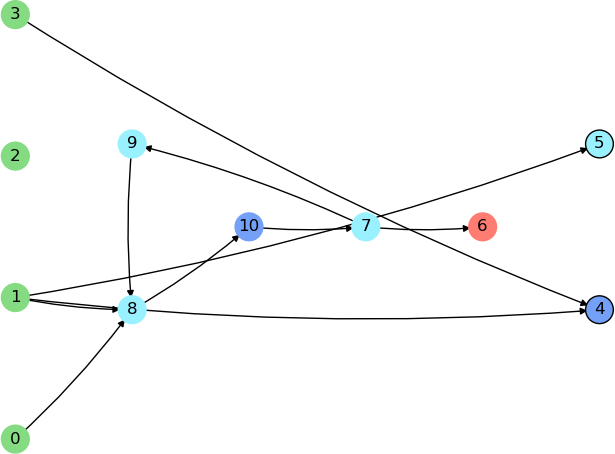}
    \caption{Numeric identifiers} \label{fig:1d-topology-a}
  \end{subfigure} \hspace*{\fill} 
  \begin{subfigure}{0.49\textwidth}
    \includegraphics[width=\linewidth]{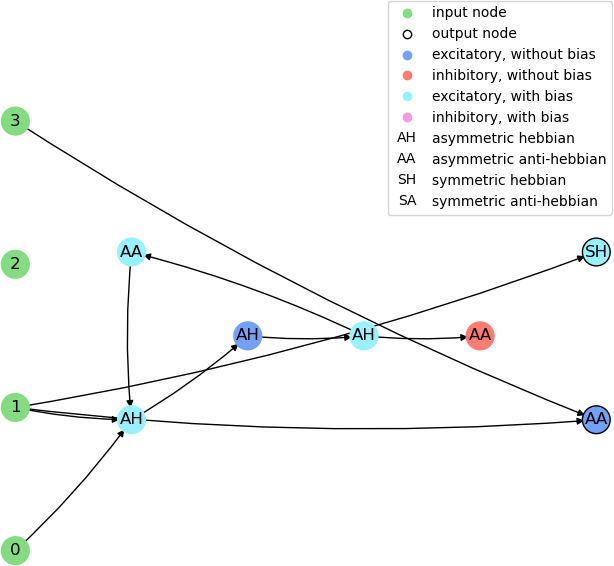}
    \caption{Learning rules} \label{fig:1d-topology-b}
  \end{subfigure} 
\caption{\label{fig:1d-topology} Illustration of the network topology of the highest accuracy agent in the food foraging task. The one-hot encoded input sample goes into nodes 0 and 1, the reward signal in node 2, and the penalty signal goes into node 3. Node 4 is the output for the `eat' actuator and node 5 is the output for the `avoid' actuator.}
\end{figure}

\begin{figure*}[htb]
\includegraphics[width=\textwidth]{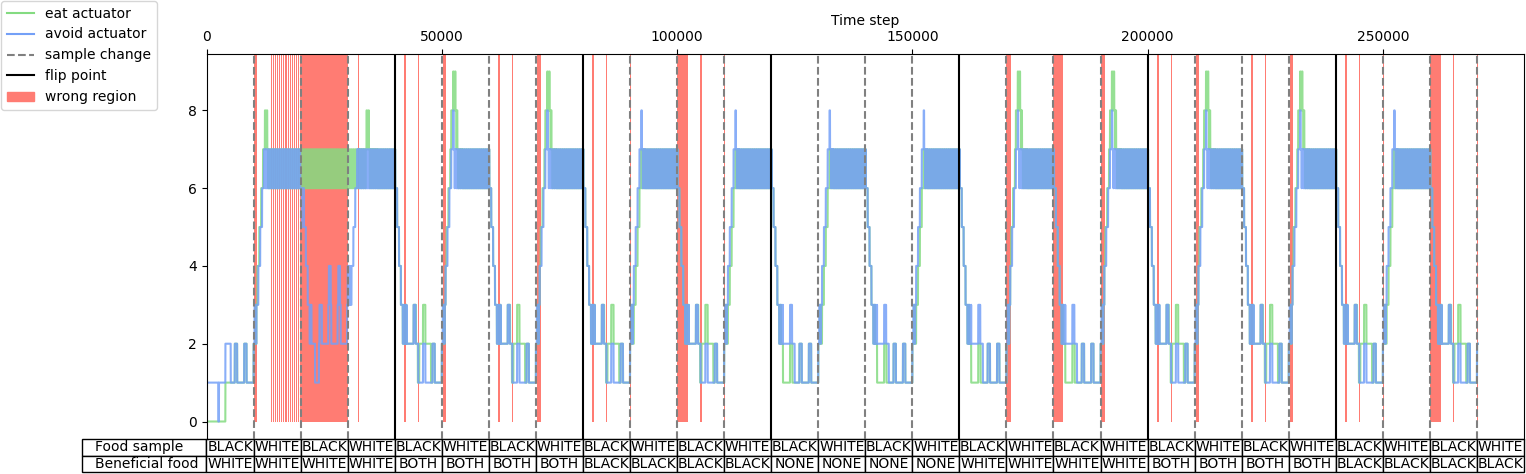}
\centering
\caption{Visualization of the spike count of the agent's actuators during test simulation \#1 for the food foraging task.}
\label{fig:food_actuator}
\end{figure*}

Fig.~\ref{fig:food_res} shows the evolution history of the food foraging task. The average fitness score has a slight increase, but the maximum fitness score does not follow this trend. The accuracy and end-of-sample accuracy have high variation with their maximum values, but they consist of high accuracies. Moreover, some early generations register 100\% end-of-sample accuracy. The three measurements do not improve through the generations. However, good solutions are already found in the first generation. Therefore, this is an easy task that requires a small SNN. For test simulations, we select the individual with the highest accuracy, which is found in generation number 34 and has an accuracy of 89.8\%. Its fitness score is 0.541395 and its end-of-sample accuracy is 100\%. Its topology is shown in Fig.~\ref{fig:1d-topology}. Paying attention to this topology, the hidden nodes are not needed. They form a loop that does not connect with the output nodes. The topology summarizes in one of the one-hot encoded input nodes (node 1) connecting to the two output nodes. Then, the node with the penalty signal (node 3) connects only with the node for the `eat' actuator (node 4). The behavior of the network is illustrated in Fig.~\ref{fig:food_actuator}. The topology of the network indicates that the two output neurons have the same data input from node 1, but the neuron for `avoid' action has a bias, which gives it a small excitatory current. If `avoid' is the wrong action, the penalty input signal from node 3 excites the output neuron for the `eat' action. This is how the spiking neural network decides the actions from "understanding" the feedback of the environment given by the penalty input signal. The result of the ten test simulations is presented in Tab.~\ref{tab:1d-sim}.

\begin{table}[!ht]
    \centering
    \begin{tabularx}{0.76\textwidth} { |c|c|c|l|X| }
    \hline
    \multicolumn{5}{|c|}{Food Foraging Test Simulations} \\ \hline
    \# & Acc. & EOS Acc. & Input Order & Environment Order \\ \hline 
    1 & 88.0\% & 92.6\% & black, white & white, both, black, none \\ \hline
    2 & 90.6\% & 100\% & white, black & white, none, both, black \\ \hline
    3 & 91.3\% & 100\% & black, white & white, both, none, black \\ \hline
    4 & 85.4\% & 92.3\% & white, black & white, black, both, none \\ \hline
    5 & 89.5\% & 96.3\% & white, black & both, none, white, black \\ \hline
    6 & 89.2\% & 100.0\% & black, white & both, white, black, none \\ \hline
    7 & 87.7\% & 92.6\% & black, white & white, black, none, both \\ \hline
    8 & 84.9\% & 92.6\% & black, white & black, both, white, none \\ \hline 
    9 & 89.8\% & 100\% & black, white & white, black, both, none \\ \hline 
    10 & 88.4\% & 92.6\% & white, black & black, none, white, both \\ \hline \hline
    Avg. & 88.4\% & 95.9\% & \multicolumn{2}{c|}{n/a} \\ \hline
    \end{tabularx}
    \caption{\label{tab:1d-sim} Test simulations of the highest accuracy agent in the food foraging experiment. "Acc." stands for accuracy and "EOS Acc." for end-of-sample accuracy.}
\end{table}

\begin{figure}[tb]
\centering
\subfloat[Fitness]{\label{fig:logic_fit}\includegraphics[width=0.31\textwidth]{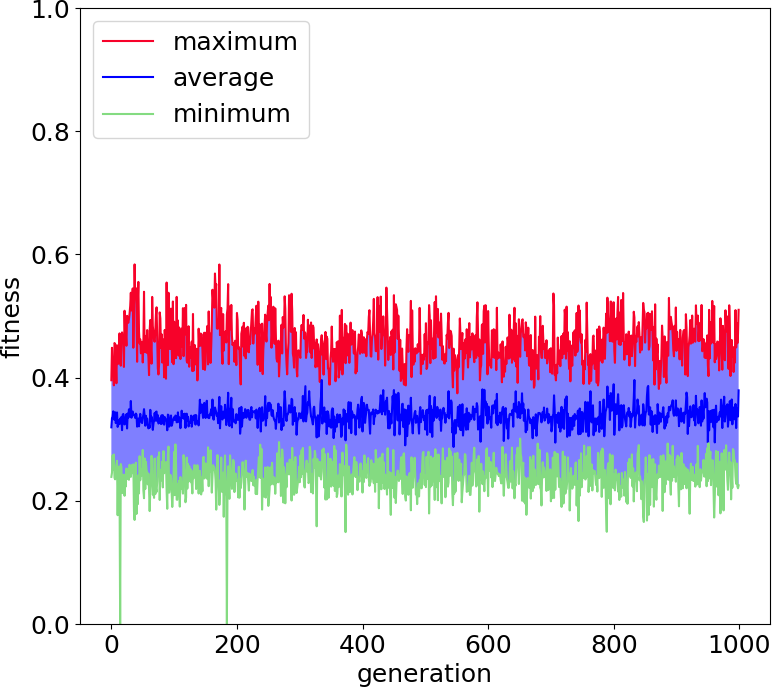}}
\hfill
\subfloat[Accuracy]{\label{fig:logic_acc}\includegraphics[width=0.31\textwidth]{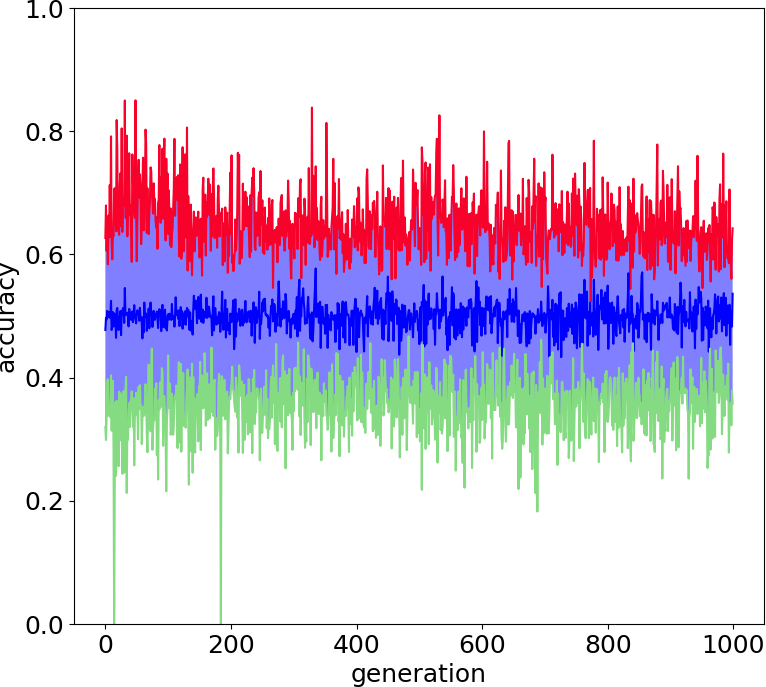}}
\hfill
\subfloat[End-of-sample accuracy]{\label{fig:logic_eos_acc}\includegraphics[width=0.31\textwidth]{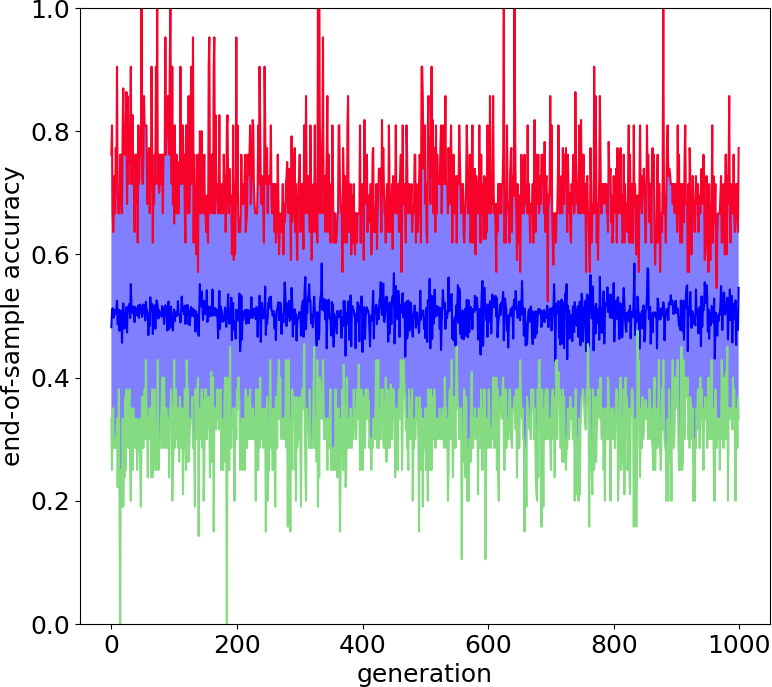}}
\caption{Evolution history of logic gate environment showing the average, minimum and maximum per generation.}
\label{fig:logic_res}
\end{figure}

\begin{figure}[!htb]
\centering
  \begin{subfigure}{0.49\textwidth}
    \includegraphics[width=\linewidth]{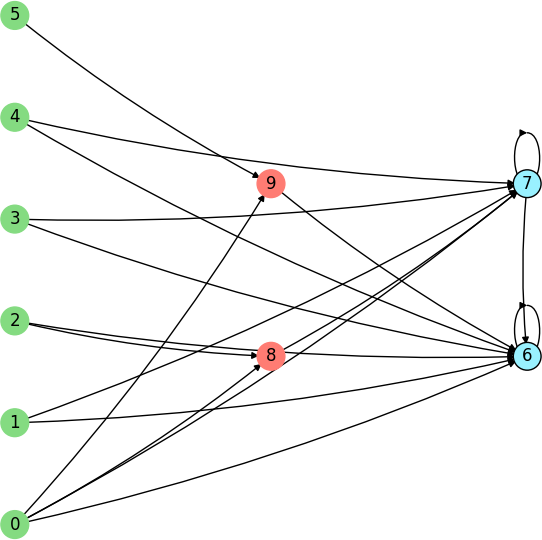}
    \caption{Numeric identifiers} \label{fig:2d-topology-a}
  \end{subfigure}\hspace*{\fill} 
  \begin{subfigure}{0.49\textwidth}
    \includegraphics[width=\linewidth]{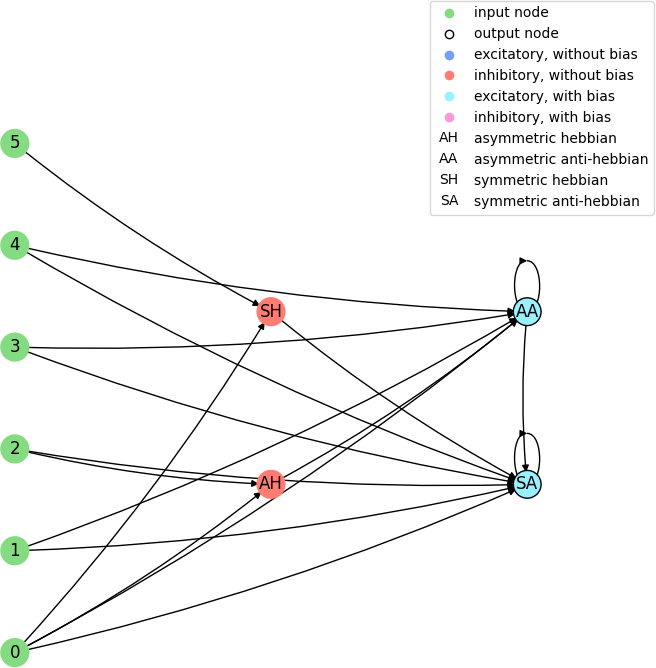}
    \caption{Learning rules} \label{fig:2d-topology-b}
  \end{subfigure} 
\caption{\label{fig:2d-topology} Illustration of the network topology of the highest training accuracy agent in the logic gate task. The one-hot encoded input sample `A' goes into nodes 0 and 1, the one-hot encoded input sample `B' goes into nodes 2 and 3, the reward signal goes into node 4, and the penalty signal into node 5. Node 6 is the output for the `0' actuator and node 7 is the output for the `1' actuator.}
\end{figure}

\begin{figure*}[htb]
\includegraphics[width=\textwidth]{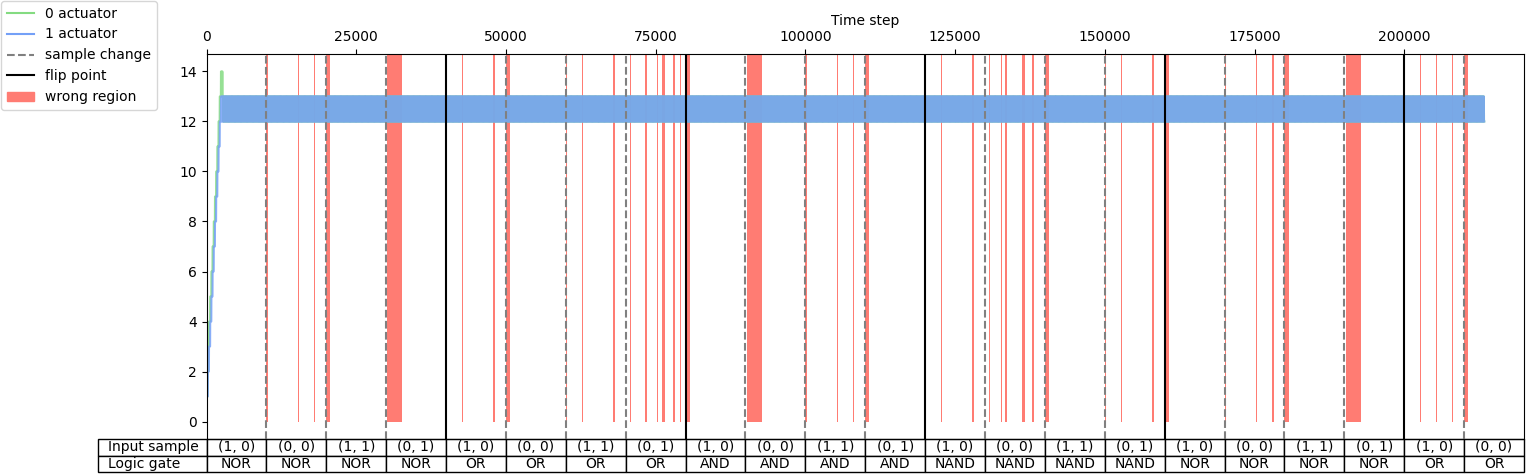}
\centering
\caption{Visualization of the spike count of the agent's actuators during test simulation \#1 for the logic gate task.}
\label{fig:logic_actuator}
\end{figure*}

Fig.~\ref{fig:logic_res} shows the training results of the logic gate task and it includes the test of the maximum individual of the measurement in every generation. The fitness score, accuracy, and end-of-sample accuracy maintain average values with high variation. However, the evolution of the agents in the logic gate task is similar to the one in the food foraging. The early generations already contain good spiking neural networks for the task. The best-performing agent is selected from the accuracy measurement. This individual is in generation 48 and has an accuracy of 85.0\%. Its fitness score is 0.4421625 and its end-of-sample accuracy is 100\%. The topology of this spiking neural network is shown in Fig.~\ref{fig:2d-topology}. Its behavior is shown in Fig.~\ref{fig:logic_actuator}. Even though we have trained with a "confidence" factor in the fitness function, the spike counts are still with almost the same values. Tab.~\ref{tab:2d-sim} contains the accuracy and end-of-sample accuracy of ten test simulations, which indicates that the SNN can be general to reproduce the behavior of logic gates without being trained to them.



\begin{table}[ht]
    \centering
    \begin{tabularx}{0.9\textwidth} { |c|c|c|l|X| }
    \hline
    \multicolumn{5}{|c|}{Logic Gate Test Simulations} \\ \hline
    \# & Acc. & EOS Acc. & Input Order (A, B) & Environment Order \\ \hline 
    1 & 89.8\% & 100\% & (1, 0), (0, 0), (1, 1), (0, 1) & NOR, OR, AND, NAND \\ \hline
    2 & 85.2\% & 95.2\% & (1, 1), (1, 0), (0, 0), (0, 1) & OR, NOR, NAND, AND \\ \hline
    3 & 86.0\% & 100\% & (1, 0), (1, 1), (0, 1), (0, 0) & NOR, OR, AND, NAND \\ \hline
    4 & 85.9\% & 95.2\% & (0, 0), (1, 1), (0, 1), (1, 0) & NAND, AND, OR, NOR \\ \hline
    5 & 79.9\% & 85.7\% & (0, 0), (0, 1), (1, 0), (1, 1) & NAND, AND, NOR, OR \\ \hline
    6 & 88.8\% & 100\% & (1, 0), (0, 0), (1, 1), (0, 1) & AND, NAND, OR, NOR \\ \hline
    7 & 85.1\% & 90.5\% & (0, 0), (1, 1), (1, 0), (0, 1) & OR, NOR, NAND, AND \\ \hline
    8 & 84.8\% & 90.5\% & (1, 1), (0, 1), (0, 0), (1, 0) & NOR, NAND, OR, AND \\ \hline
    9 & 83.7\% & 85.7\% & (0, 0), (1, 0), (0, 1), (1, 1) & NAND, NOR, OR, AND \\ \hline
    10 & 88.5\% & 100\% & (1, 1), (1, 0), (0, 0), (0, 1) & NOR, AND, OR, NAND \\ \hline \hline
    Avg. & 85.7\% & 94.2\% & \multicolumn{2}{c|}{n/a} \\ \hline
    \end{tabularx}
    \caption{\label{tab:2d-sim} Test simulations of the highest training accuracy agent in the logic gate experiment. "Acc." stands for accuracy and "EOS Acc." for end-of-sample accuracy.}
\end{table}

\begin{figure}[tb]
\centering
\includegraphics[width=0.51\textwidth]{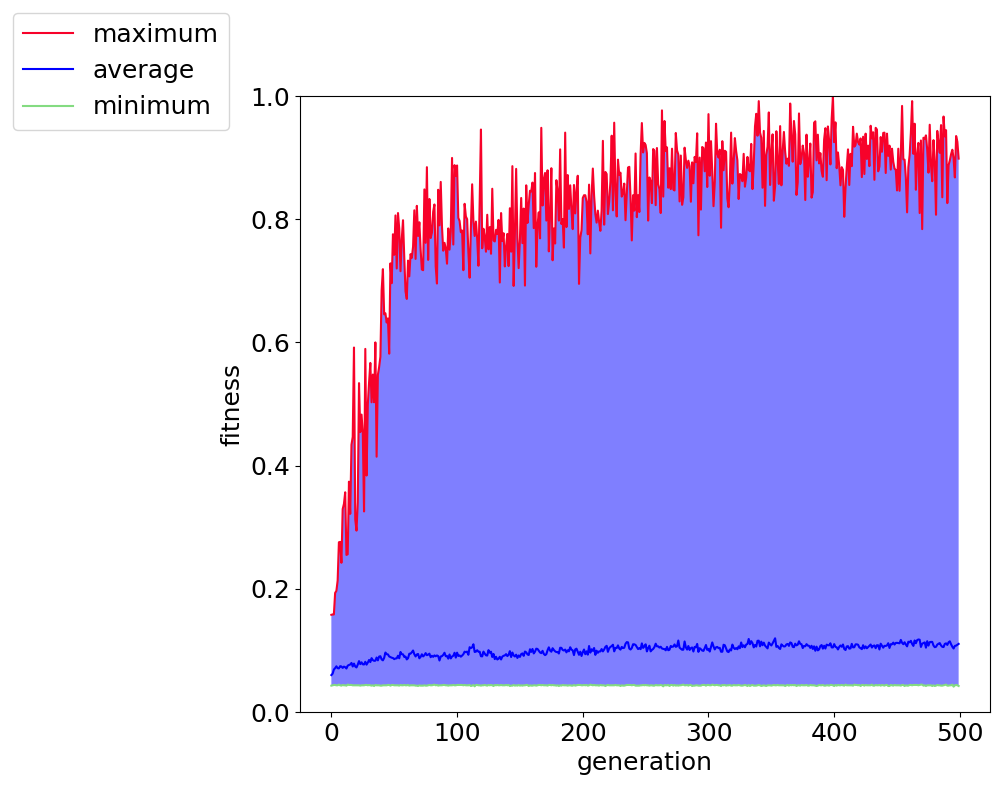}
\caption{Fitness history of cart-pole balancing environment showing the average, minimum and maximum per generation.}
\label{fig:cart_res}
\end{figure}

Fig.~\ref{fig:cart_res} shows the fitness score history through the evolution for the cart-pole balancing task. This task is the one with the highest difficulty to find a good genome for the adaptive spiking neural network. It can be noted that the fitness score improves through the generations. The maximum fitness score in a generation goes from around 0.16 in the first generation to 0.99944 in generation number 399. Such an individual is the one selected for the test simulations. Its topology is illustrated in Fig.~\ref{fig:2d-topology} and the spike counts of the actuators for `left' and `right' actions are shown in Fig.~\ref{fig:logic_actuator}. The spiking neural network has no hidden neurons. Therefore, the SNN works as an input selection for the output neurons. The result of the ten test simulations is presented in Tab.~\ref{tab:cart-sim}. When the pole is balanced for more than 100 iterations, the controller is considered successful.


\begin{figure}[!htb]
\centering
  \begin{subfigure}{0.49\textwidth}
    \includegraphics[width=\linewidth]{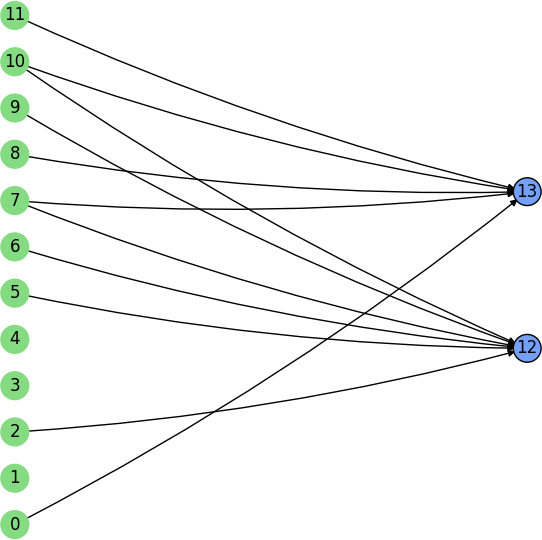}
    \caption{Numeric identifiers} \label{fig:cart-topology-a}
  \end{subfigure}\hspace*{\fill} 
  \begin{subfigure}{0.49\textwidth}
    \includegraphics[width=\linewidth]{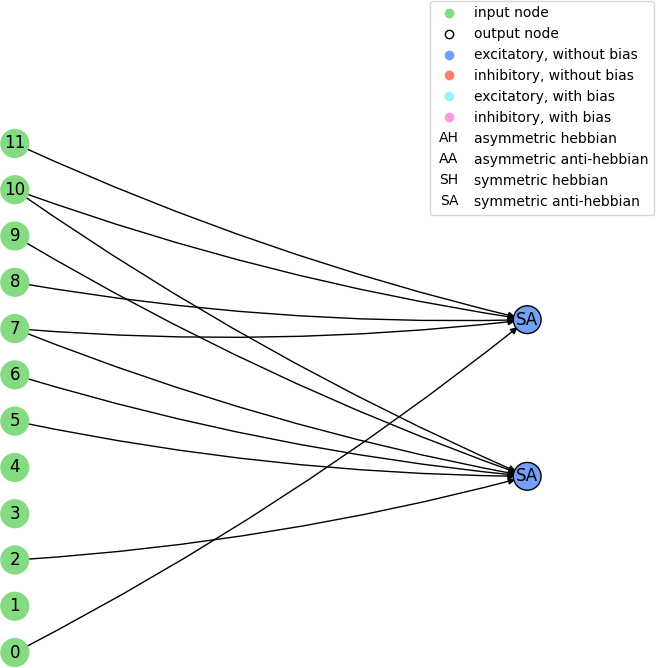}
    \caption{Learning rules} \label{fig:cart-topology-b}
  \end{subfigure} 
\caption{\label{fig:cart-topology} Illustration of the network topology of the highest training fitness agent in the cart-pole balancing experiment. The 3-tuple of input nodes (0, 1, 2) gets the converted firing rate from the observation of the cart position, (3, 4, 5) from the cart velocity, (6, 7, 8) from the pole angle, and (9, 10, 11) from the pole angular velocity. Node 12 is the output for the `left' action and node 13 is the output for the `right' action.}
\end{figure}

\begin{figure*}[htb]
\includegraphics[width=\textwidth]{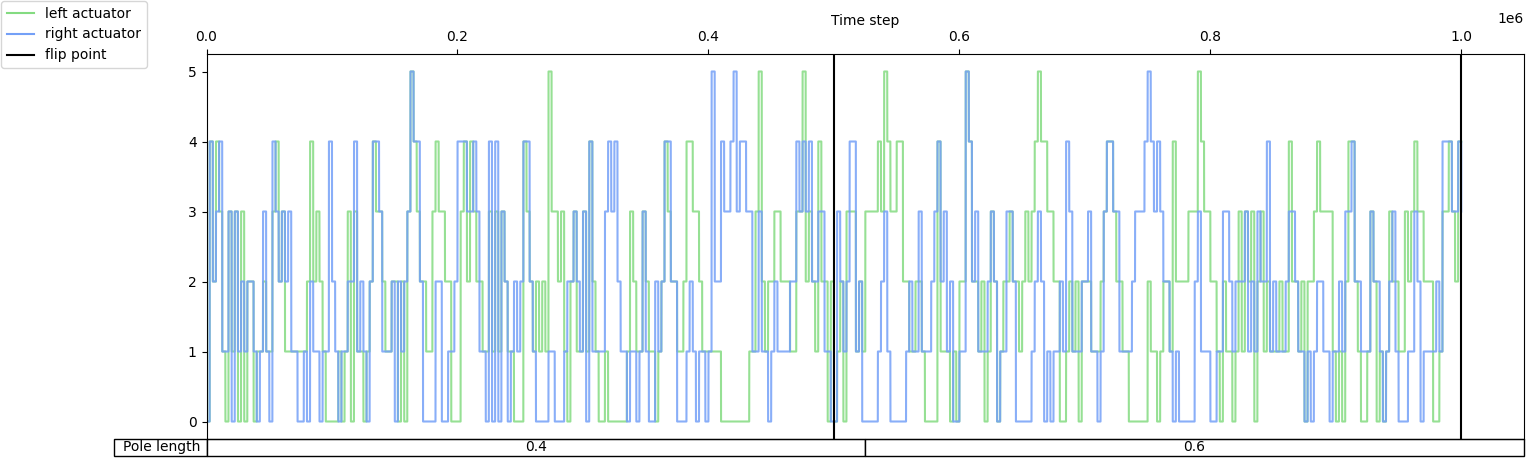}
\centering
\caption{Visualization of the spike count of the agent's actuators during test simulation \#1 for the cart-pole balancing task.}
\label{fig:cart_actuator}
\end{figure*}

\begin{table}[ht]
    \centering
    \begin{tabularx}{0.7\textwidth}{|c|c|c|c|Y|}
    \hline
    \multicolumn{5}{|c|}{Cart-pole Balancing Test Simulations} \\ \hline
    \# & Fitness & \# Steps 0.4 & \# Steps 0.6 &  Environment Order \\ \hline 
    1 & 1.000 & 200 & 200 & 0.4, 0.6 \\ \hline 
    2 & 1.000 & 200 & 200 & 0.4, 0.6 \\ \hline
    3 & 1.000 & 200 & 200 & 0.6, 0.4 \\ \hline
    4 & 0.943 & 200 & 177 & 0.4, 0.6 \\ \hline
    5 & 0.800 & 154 & 166 & 0.6, 0.4 \\ \hline
    6 & 0.792 & 179 & 138 & 0.4, 0.6 \\ \hline 
    7 & 0.835 & 200 & 134 & 0.6, 0.4 \\ \hline 
    8 & 0.845 & 200 & 138 & 0.4, 0.6 \\ \hline
    9 & 0.873 & 200 & 149 & 0.6, 0.4 \\ \hline 
    10 & 0.720 & 88 & 200 & 0.6, 0.4 \\ \hline \hline
    Avg. & 0.874 & 178.3 & 171.5 & n/a \\ \hline
    \end{tabularx}
    \caption{\label{tab:cart-sim} Test simulations of the highest fitness agent in the cart-pole balancing experiment.}
\end{table}




\section{Discussion and conclusion}
\label{sec:discussion}

We successfully solved all three presented tasks with the NAGI framework. The spiking neural networks found showed generality to the binary classification tasks, even to unseen conditions in the case of the emulation of logic gates. The neuroevolution produced rather simple topologies for the SNNs. We infer that binary classification is easy due to the binary performance feedback. For further research, multi-class classification is considered.

The cart-pole balancing task was successfully solved without any hidden neurons. The conversion of one observation into three input neurons is used to avoid the requirement of weight fine-tuning due to small differences in firing rate and also to the assumption that Hebbian plasticity works better with binary data (active and inactive) \citep{pontes2019bidirectional}. With such a conversion, the SNN became an input selection.

The topologies for the three tasks caught our attention because almost all output excitatory neurons were anti-Hebbian, and the two inhibitory hidden neurons in the logic gate solution have Hebbian neuroplasticity. Our initial hypotheses were that excitatory neurons mainly have Hebbian learning rules, and inhibitory neurons are anti-Hebbian. That was the reason for having different probabilities for anti-Hebbian and Hebbian learning rules depending on the type of the neurotransmitter when adding a new neuron through mutation.

Even though there is elitism, the performance measurements are unstable through generations. This is a demonstration of the randomness in the initialization of the weights, and input and environment order. This can be perceived in the results of the ten test simulations of the three tasks.

For future work, we plan to attempt more challenging tasks. If there is a failure in executing the task, the constraints imposed on NAGI can be eased. A major constraint is that one neuron has one plasticity rule for all dendrites. Maybe its removal can simplify issues in difficult tasks. This constraint was intended to reduce the dimensionality of the search space in the neuroevolution and an assumption that the dendrites in the same neuron adapt under one learning rule. This modification is also aligned with the work of \citet{najarro2020meta}, which has meta-learning properties for more difficult control tasks than the cart-pole balancing, such as top-down car racing and quadruped walk. Another opportunity is the addition of curriculum learning \citep{bengio2009curriculum,narvekar2020curriculum} for increasing the complexity of the task while the agent becomes better over the generations.

\section*{Acknowledgment}

This work was partially funded by the Norwegian Research Council (NFR) through their IKTPLUSS research and innovation action under the project Socrates (grant agreement 270961).

\FloatBarrier
\bibliographystyle{unsrtnat}
\bibliography{references}

\end{document}